%% file: main.tex
\documentclass[letterpaper, 10 pt, journal, twoside]{IEEEtran}  % Use this command for final RAL version

\usepackage{amsmath} % assumes amsmath package installed
\usepackage{amssymb}  % assumes amsmath package installed

\makeatletter
\let\NAT@parse\undefined
\makeatother
\usepackage[numbers,sort]{natbib}

\usepackage[breaklinks,colorlinks=true,citecolor=green]{hyperref}
\usepackage{booktabs}
\usepackage{graphicx}

\usepackage{xcolor}
\usepackage{algorithm}
\usepackage{algorithmic}

\usepackage[capitalize]{cleveref}
\crefname{section}{Sec.}{Secs.}
\crefname{table}{Tab.}{Tabs.}
\crefname{algorithm}{Alg.}{Algs.}

\title{
ImplicitRDP: An End-to-End Visual-Force Diffusion Policy\\ with Structural Slow-Fast Learning
}

\author{Wendi~Chen$^{1,2}$, Han~Xue$^{1}$, Yi~Wang$^{1,2}$, Fangyuan~Zhou$^{1,2}$, Jun~Lv$^{1,3}$, Yang~Jin$^{1}$, Shirun~Tang$^{3}$,\\ Chuan~Wen$^{1,\dagger}$, and Cewu~Lu$^{1,2,3,\dagger}$%
\thanks{Manuscript received: December 4, 2025; Revised: May 23, 2026; Accepted: June 20, 2026.}%
\thanks{This paper was recommended for publication by Editor Wei Pan upon evaluation of the Associate Editor and Reviewers' comments. 
This work was supported by Fundamental and Interdisciplinary Disciplines Breakthrough Plan of the Ministry of Education of China, Science and Technology Major Project of Jiangsu Province (No. BG2024041), Shanghai Artificial Intelligence Laboratory, and XPLORER PRIZE grants.}%
\thanks{$^{1}$Wendi Chen, Han Xue, Yi Wang, Fangyuan Zhou, Jun Lv, Yang Jin, Chuan Wen, and Cewu Lu are with Shanghai Jiao Tong University, Shanghai, China.}%
\thanks{$^{2}$Wendi Chen, Yi Wang, Fangyuan Zhou, and Cewu Lu are with Shanghai Innovation Institute, Shanghai, China.}%
\thanks{$^{3}$Jun Lv, Shirun Tang, and Cewu Lu are with Noematrix Ltd., Shanghai, China.}%
\thanks{$^\dagger$Corresponding authors: Chuan Wen ({\tt\footnotesize wenchuan@sjtu.edu.cn}) and Cewu Lu ({\tt\footnotesize lucewu@sjtu.edu.cn}).}%
\thanks{Digital Object Identifier (DOI): see top of this page.}%
}

\begin{document}

\maketitle

%%%%%%%%%%%%%%%%%%%%%%%%%%%%%%%%%%%%%%%%%%%%%%%%%%%%%%%%%%%%%%%%%%%%%%%%%%%%%%%%
\begin{abstract}
Human-level contact-rich manipulation relies on the distinct roles of two key modalities: vision provides spatially rich but temporally slow global context, while force sensing captures rapid local contact dynamics.
Integrating these signals is challenging due to their fundamental frequency and informational disparities.
In this work, we propose ImplicitRDP, a unified end-to-end visual-force diffusion policy that integrates visual planning and reactive force control within a single network.
We introduce \textit{Structural Slow-Fast Learning}, a mechanism utilizing causal attention to simultaneously process asynchronous visual and force tokens, 
allowing the policy to perform rapid force control at the action rate while maintaining the temporal coherence of action chunks.
Furthermore, to mitigate modality collapse where end-to-end models fail to adjust the weights across different modalities,
we propose \textit{Virtual-target-based Representation Regularization}.
This auxiliary objective maps force feedback into the same space as the action, providing a stronger, physics-grounded learning signal than raw force prediction.
Extensive experiments on contact-rich tasks demonstrate that ImplicitRDP significantly outperforms both vision-only and hierarchical baselines, achieving superior reactivity and success rates with a streamlined training pipeline. Code and videos are available at \href{https://implicit-rdp.github.io}{implicit-rdp.github.io}.

\end{abstract}


%%%%%%%%%%%%%%%%%%%%%%%%%%%%%%%%%%%%%%%%%%%%%%%%%%%%%%%%%%%%%%%%%%%%%%%%%%%%%%%%

\input{sections/1_introduction}
\input{sections/2_related_work}
\input{sections/3_methodology}
\input{sections/4_experiments}
\input{sections/5_conclusion}

%%%%%%%%%%%%%%%%%%%%%%%%%%%%%%%%%%%%%%%%%%%%%%%%%%%%%%%%%%%%%%%%%%%%%%%%%%%%%%%%

\section*{Acknowledgment}
We would like to thank Wenye Yu, Yanwen Zou, Lifeng Zhuo, Yuan Fang, and Junjie Fang for insightful discussions.

\bibliographystyle{IEEEtran}
\bibliography{references}

\end{document}

%% file: sections/1_introduction.tex
% !TEX root = ../main.tex
\section{Introduction}

\IEEEPARstart{H}{uman} manipulation is fundamentally a multi-modal endeavor.
While we rely on vision to plan trajectories and locate objects globally, we depend heavily on force sensing to manage the precise dynamics of interaction locally.
Especially in contact-rich tasks, vision alone is far from sufficient, due to severe occlusions and the inherent ambiguity of contact states.
To replicate this human-level capability, robot learning must effectively integrate both vision and force modalities.

While imitation-based policy learning has achieved remarkable success in recent years \cite{dp, act, pi0, pi0.5, gen0}, integrating force feedback into existing policy architectures remains a non-trivial challenge due to the fundamental disparities between the two signals.
\textbf{Vision} provides the ``global context" of the environment, which is spatially rich but can be temporally slow ($1\sim2$ Hz) under the action chunking setting.
\textbf{Force}, conversely, reflects the ``local reality" of contact, changing rapidly during interaction and therefore requiring timely feedback at the action rate ($10$ Hz).

To bridge this frequency gap between the two modalities, recent approaches such as Reactive Diffusion Policy (RDP) \cite{rdp} employ a hierarchical ``slow-fast" framework.
These systems decouple the problem: a slow policy process vision to generate latent actions, which guide a fast policy processing force to predict reactive real actions in a closed loop.
While intuitive, this explicit separation introduces some flaws.
First, it creates an information bottleneck: the fast policy is effectively ``blind" to the spatial geometry, relying solely on compressed latent actions.
Second, it suffers from modal conflict: if the slow policy makes a semantic error, the fast policy lacks the context to correct the plan, and the entire system becomes highly fragile due to compounding errors between modules.
Finally, the ``hand-over" between vision and force is rigidly hand-designed, limiting the model's scalability to learn complex behaviors.

\begin{figure}[!tbp]
    \centering
    \includegraphics[width=\linewidth]{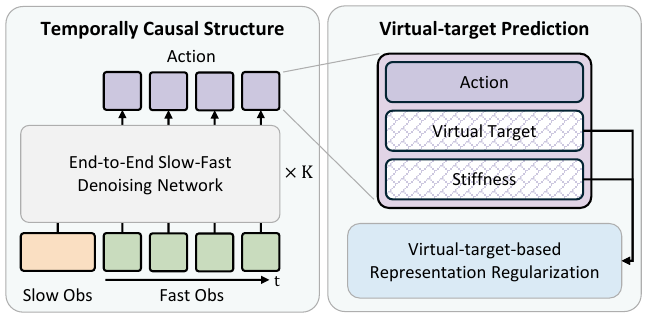}
    \vspace{-6mm}
    \caption{\textbf{Structural Slow-Fast Learning:} we leverage a temporally causal structure to enable end-to-end closed-loop force-based control within an action chunk.
    \textbf{Virtual-target-based Representation Regularization}: we further incorporate virtual target prediction as an auxiliary task to prevent modality collapse.}
    \label{fig:teaser}
    \vspace{-4mm}
\end{figure}

In this paper, we propose \textbf{ImplicitRDP} (see \cref{fig:teaser}), an end-to-end policy that integrates these disparate signals within a unified Transformer architecture.
Instead of enforcing a rigid hierarchy, we treat multi-modal control as a sequence modeling problem.
We concatenate low-frequency visual tokens and faster force tokens into a unified sequence, using causal cross-attention to structure the interaction between actions and different modalities, which is termed as \textit{Structural Slow-fast Learning}.
This design eliminates the information bottleneck by allowing action tokens to directly attend to raw visual and force tokens simultaneously,
therefore unlocking full spatial awareness during contact and enabling geometry-aware force control.
Moreover, we show that with a consistent inference mechanism, a diffusion model equipped with a causal structure can also perform closed-loop control over force signals.

While the structural slow-fast policy achieves scalable end-to-end training, we observe a modality collapse problem in this framework,
where the model becomes overly dependent on a single modality and fails to adaptively adjust the weighting across different modalities \cite{factr}.
To counter this, we introduce \textit{Virtual-Target-based Representation Regularization}.
Rather than predicting future end-effector force, we train the model to predict a ``virtual target" that implies the desired force profile.
By mapping force requirements into the Cartesian action space, we provide a stronger, more actionable learning signal that forces the network to adaptively attend to force feedback.

We evaluate ImplicitRDP on representative contact-rich tasks, including box flipping, switch toggling, and hole inserting. Experiments demonstrate that our end-to-end approach outperforms explicitly hierarchical baselines, offering a simpler training pipeline while achieving superior reactivity and robustness.

Overall, this paper makes the following contributions:
\begin{itemize}
  \item We propose ImplicitRDP, an end-to-end visual-force policy with structural slow-fast learning that simultaneously processes slow and fast observations while realizing rapid force control at the action rate.
  \item We introduce an auxiliary task based on virtual-target prediction that encourages the policy to adaptively adjust weights of different modalities. It maps desired forces into the Cartesian action space and appropriately weights losses according to force magnitudes, leading to more effective guidance than conventional force prediction.
  \item Extensive experiments on three representative contact-rich tasks demonstrate that ImplicitRDP achieves higher performance than baseline methods and provides a more streamlined and unified training framework as well.
\end{itemize}

Code and videos are available at \href{https://implicit-rdp.github.io}{implicit-rdp.github.io}.

%% file: sections/2_related_work.tex
% !TEX root = ../main.tex
\section{Related Works}

\subsection{Imitation Learning with Force Input}
Imitation Learning (IL) has emerged as a dominant paradigm in robotic manipulation \cite{dp, act}.
Notable works \cite{pi0, pi0.5, gen0} have demonstrated exceptional scalability when utilizing visual observations as input, 
successfully tackling complex challenges ranging from deformable object manipulation to long-horizon assembly.

Recent research \cite{tacdiffusion, acp, forcemimic, dexforce, cr_dagger, forcevla} has begun to integrate force and torque measurements as additional modalities within the IL framework, 
which aims to enhance the model's understanding of contact states and exerted forces, and improve performance in contact-rich scenarios.
Most of these works incorporate force/torque signals from the robot's Tool Center Point (TCP) directly into the policy input.
However, there is a critical limitation in these approaches. Due to action chunking \cite{dp, act}, the control within each chunk remains effectively open-loop. 
This prevents the system from reacting to force feedback within each action chunk at the action rate,
which consequently constrains performance in contact-rich tasks.
To address this, Reactive Diffusion Policy (RDP) \cite{rdp} proposed a hierarchical slow-fast architecture to achieve closed-loop control based on force signals.
It utilizes a slow network to predict latent action and a fast network to decode the latent combined with the latest force signals into executable actions.

Despite its effectiveness, the two-stage design of RDP introduces complexity in training and hyperparameter tuning and limits potential scalability.
In this work, we propose ImplicitRDP,  which achieves similar force-based closed-loop control but within a unified framework.
Inspired by the success of causal modeling in domains such as large language models \cite{gpt4, qwen3} and video generation\cite{causvid}, we implement a structural slow-fast learning mechanism via causal attention.
This allows rapid force control at the action rate while preserving end-to-end training.

\subsection{Mitigate Modality Collapse}
While RDP enforces attention to different modalities through its hierarchical architecture,
standard end-to-end networks often struggle with modality collapse where models are unable to flexibly adjust the weights across multiple modalities.

To mitigate this, FACTR \cite{factr} introduced a curriculum learning strategy that blurs visual inputs during the early stages of training,
guiding the network to autonomously learn how to weight different modalities.
However, this approach adds training complexity and reduces generalizability across different tasks.

Alternatively, introducing future prediction as representation regularization has shown promise in robotic manipulation \cite{gr1, seer, vpp, uva, uwm}.
These works demonstrate that using future observation prediction as an auxiliary task significantly enhances policy robustness and representation quality.
Building on this insight, recent work has applied future prediction to policies with force input.
TA-VLA \cite{ta_vla} employs future torque prediction as an additional objective,
finding that it encourages the model to learn physically grounded internal representations and improves manipulation performance.

In this work, we also leverage the paradigm of future prediction but propose a novel target. Inspired by classical compliance controllers, we utilize the virtual target, which is calculated via adaptive compliance parameters, as the prediction objective.
Unlike raw force/torque prediction as in TA-VLA, the virtual target is directly aligned with the Cartesian action space used by the policy while assigning magnitude-dependent weights to force signals.
Compared with compliance-control methods such as ACP \cite{acp}, our method only uses adaptive compliance to construct an auxiliary target for position-controlled imitation learning.
We demonstrate that this approach provides tighter representation regularization, facilitating more effective utilization of different modalities.

%% file: sections/3_methodology.tex
% !TEX root = ../main.tex
\section{Methodology}

We propose ImplicitRDP, an end-to-end framework that unifies visual planning and rapid force control at the action rate.
In this section, we first introduce the preliminaries of standard Diffusion Policy  \cite{dp} in \cref{sec:preliminary}. 
Then, we detail the implementation of \textit{structural slow-fast learning (SSL)} in \cref{sec:structural_slow_fast} and the \textit{virtual-target-based representation regularization (VRR)} in \cref{sec:virtual_target}.
Finally, we discuss critical implementation details in \cref{sec:implementation}.

\subsection{Preliminary: Diffusion Policy}
\label{sec:preliminary}
Diffusion Policy (DP) formulates robotic control as a conditional generative modeling problem.
Specifically, it learns the conditional distribution $p(\mathbf{A}_t | \mathbf{O}_t)$ of action sequences $\mathbf{A}_t\triangleq\{a_{t-h_o+1}, \dots, a_{t-h_o+h_a}\}$ with horizon ${h_a}$ given observations $\mathbf{O}_t\triangleq\{o_{t-h_o+1}, \dots, o_{t}\}$ with horizon $h_o$ at time ${t}$.
During the training process, a Gaussian noise $\epsilon^k$ is added to the ground-truth action $\mathbf{A}^0_t$ via \cref{eq:diffuse},
where $k$ is the diffusion step and $\bar{\alpha}_k$ is the noise schedule.
\begin{equation}
\label{eq:diffuse}
\mathbf{A}^k_t = \sqrt{\bar{\alpha}_k}\mathbf{A}^0_t + \sqrt{1-\bar{\alpha}_k}\epsilon^k.
\end{equation}
Then, a noise prediction network $\epsilon_\theta$ is trained to predict the noise given the noisy action $\mathbf{A}^k_t$ and observation $\mathbf{O}_t$.
The DDPM \cite{ddpm} loss is defined in \cref{eq:dp_loss}.
\begin{equation}
\label{eq:dp_loss}
\mathcal{L}_{\epsilon} = \mathbb{E}_{k, \epsilon^k, (\mathbf{O}_t, \mathbf{A}^0_t)} [\| \epsilon^k - \epsilon_\theta(\mathbf{O}_t, \mathbf{A}^k_t, k) \|^2].
\end{equation}
During inference, starting from a Gaussian noise $\mathbf{A}^K_t \sim \mathcal{N}(0, \mathbf{I})$, the noise prediction network $\epsilon_\theta$ iteratively denoises it through Stochastic Langevin Dynamics \cite{stochastic_langevin_dynamics} to generate the predicted action chunk $\mathbf{\hat{A}}^0_t$.
Since DP typically uses Receding Horizon Control (RHC) to execute actions, 
the execution within each chunk keeps open-loop when new force observations are available,
limiting its reactivity in contact-rich tasks.

\subsection{Structural Slow-Fast Learning}
\label{sec:structural_slow_fast}
To enable rapid force control at the action rate within action chunking,
we introduce \textit{structural slow-fast learning}.
Unlike RDP \cite{rdp}, which employs a two-stage slow-fast architecture,
structural slow-fast learning realizes end-to-end action modeling with variant-frequency observations through temporally causal structure and consistent inference mechanism.

\begin{figure}[htbp]
    \centering
    \includegraphics[width=0.8\linewidth]{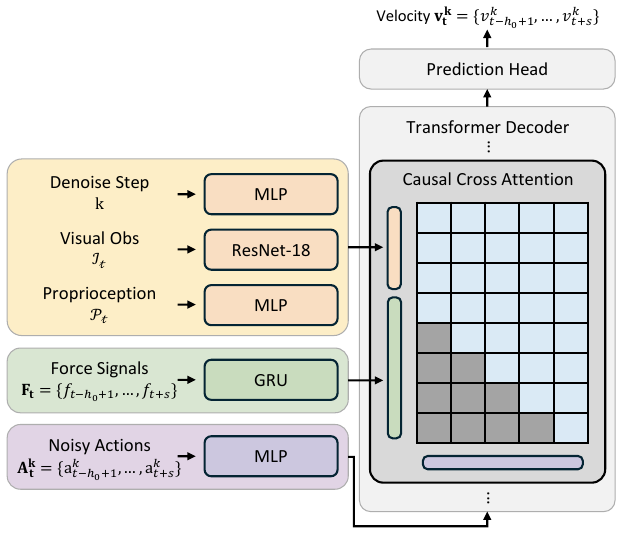}
    \vspace{-4mm}
    \caption{\textbf{Network Architecture of ImplicitRDP.} We enforce a temporally causal structure using a GRU for force signal encoding and a causal attention mask for action-force interaction, which enables structural slow-fast learning.}
    \label{fig:model_structure}
\end{figure}

\subsubsection{Temporally Causal Structure}
As shown in \cref{fig:model_structure}, we design ImplicitRDP based on the standard Transformer-based \cite{transformer} DP with ResNet-18 \cite{resnet} visual encoder.
We separate the observations into the ``slow'' part (visual observations $\mathcal{I}_t$ and proprioception $\mathcal{P}_t$) and the ``fast'' part (force signals $\mathbf{F}_t \triangleq \{f_{t-h_o+1}, \dots, f_{t-h_o+h_a}\}$). 
During training, this force sequence is aligned with the action chunk at the action-rate timestamps.
Unlike standard Transformer-based DP which directly encodes all observations and applies full cross-attention between the action tokens and observation tokens,
we treat the ``fast'' force signals as a temporal sequence aligned with the action chunk which are used for closed-loop control.
To prevent future information leakage, we modify the model structure to meet the temporal causality requirement.
First, we use GRU \cite{gru} for force encoding to preserve causality constraints.
Second, we employ a causal attention mask for the force tokens, ensuring that the prediction of action $a_{t-h_o+s} (1\le s\le h_a)$ can attend only to the current and past force tokens $\{f_{t-h_o+1}, \dots, f_{t-h_o+s}\}$.
The future force tokens, namely $\{f_{t-h_o+s+1}, \dots, f_{t-h_o+h_a}\}$, are masked out and cannot be used for predicting $a_{t-h_o+s}$.
These structural constraints allow full logged force sequences to be processed efficiently during parallel training without future force leakage.

\subsubsection{Consistent Inference Mechanism}

Owing to the temporally causal structure designed in ImplicitRDP, the model naturally supports variable-length inputs of action tokens.
We leverage this property to realize closed-loop force control in a streaming manner by continuously sampling from a noisy action sequence that extends the previous one with one additional token.
However, because of the inherent stochasticity of diffusion models, independent sampling at each step can lead to inconsistency between consecutive inferences.

To address this, we design a mechanism to guarantee consistency across multiple inference passes within a single action chunk.
We utilize the DDIM sampler \cite{ddim} and set the stochasticity parameter $\eta=0$.
Under this setting, the denoising trajectory becomes strictly deterministic given the initial noise $\mathbf{A}^K_t$.
Consequently, as described in \cref{alg:implicitrdp_inference}, we perform slow observation encoding and noise sampling only once at the beginning of a chunk.
These results are cached and reused throughout the whole chunk. 
At every control step, we fetch the fast observations of the corresponding length, encode them into fast tokens, and run the DDIM sampler using the cached noise and slow tokens combined with the updated fast tokens.
We then execute the last action of the generated sequence.
This approach enables force-based closed-loop control while at the same time maintaining the advantages brought by action chunking,
such as smoother motions and compatibility with non-Markovian behavior.

\begin{algorithm}[h]
\caption{Consistent Inference in ImplicitRDP}
\label{alg:implicitrdp_inference}
\begin{algorithmic}[1]
\REQUIRE Policy Network $\pi_\theta$, Slow Observation Encoder $\mathcal{E}_{slow}$, Fast Observation Encoder $\mathcal{E}_{fast}$, Slow Observation Horizon $h_o$, Execution Horizon $h_e$, Latency Steps $l$ (to handle slow observation encoder inference latency and other system execution delays)
\STATE Initialize global execution step $t \leftarrow 0$
\STATE \COMMENT{\textbf{slow loop}: action chunk modeling}
\WHILE{Task not done}
    \STATE \COMMENT{cache slow context and noise}
    \STATE $\mathbf{O}_{slow} \leftarrow \text{GetSlowObservation}(len=h_o)$
    \STATE $\mathbf{Z}_{slow} \leftarrow \mathcal{E}_{slow}(\mathbf{O}_{slow})$
    \STATE $\mathbf{A}^K \leftarrow \mathcal{N}(0, \mathbf{I})$
    
    \STATE \COMMENT{\textbf{fast loop}: closed-loop control within the horizon where $i$ indexes the current step within the chunk}
    \FOR{$i \leftarrow 0$ \TO $h_e - 1$}
        \STATE \COMMENT{update fast context}
        \STATE $\mathbf{O}_{fast} \leftarrow \text{GetFastObservation}(len=h_o+i+l)$
        \STATE $\mathbf{Z}_{fast} \leftarrow \mathcal{E}_{fast}(\mathbf{O}_{fast})$
        
        \STATE \COMMENT{get a noisy action sequence of a certain length}
        \STATE $\mathbf{A}^K_t \leftarrow \mathbf{A}^K[:i+h_o+l]$
        
        \STATE \COMMENT{consistent denoising with cache}
        \STATE $\mathbf{\hat{A}}^0_t \leftarrow \text{DDIM}(\pi_\theta, \mathbf{Z}_{slow}, \mathbf{Z}_{fast}, \mathbf{A}^K_t, \eta=0)$

        \STATE \COMMENT{execute the current step action}
        \STATE $a_t \leftarrow \mathbf{\hat{A}}^0_t[-1]$
        \STATE \text{Execute} $a_t$

        \STATE $t \leftarrow t + 1$
        
        \IF{Task done}
            \STATE \textbf{break}
        \ENDIF
    \ENDFOR
\ENDWHILE
\end{algorithmic}
\end{algorithm}

\subsection{Virtual-target-based Representation Regularization}
\label{sec:virtual_target}

To prevent the end-to-end policy from relying solely on a single modality, we introduce a novel representation regularization method. Inspired by \cite{ta_vla}, which predicts future torque to enforce physical understanding, we propose predicting the \textit{virtual target}.
Unlike raw force, the virtual target resides in the same Cartesian space as the robot's action, facilitating a more unified representation learning process.

\subsubsection{Virtual Target Formulation}
The concept of the virtual target is derived from compliance control theory \cite{compliance_control}.
A standard compliance system can be modeled as a spring-mass-damper system:
\begin{equation}
\label{eq:compliance_model}
f_{ext} = M\ddot{x}_{vt} + D\dot{x}_{vt} + K(x_{vt} - x_{real}),
\end{equation}
where $f_{ext}$ denotes the external wrench, and $M$, $D$, and $K$ represent the inertia, damping, and stiffness matrices.
$x_{real}$ is the robot's actual pose, and $x_{vt}$ is the virtual target.
In the context of quasi-static manipulation, we can ignore the inertia and damping terms.
Consequently, the virtual target can be derived given the current stiffness and measured force:
\begin{equation}
\label{eq:vt_calculation}
x_{vt} = x_{real} + K^{-1}f_{ext}.
\end{equation}
The raw end-effector wrench is measured by the force/torque sensor in the TCP frame.
For the translational virtual target, we transform the translational force component into the world frame and use it in \cref{eq:vt_calculation}.
The Cartesian action is also represented in the world frame, so $x_{vt}$ and the policy action are defined in the same Cartesian space.

\subsubsection{Adaptive Stiffness Assignment}
A manipulation task may consist of multiple phases, such as approaching the object, making contact, and manipulating it.
Thus, it is inappropriate to apply the same force-based representation regularization uniformly across all phases.
Instead, we adopt the heuristic strategy from ACP \cite{acp} to assign an adaptive stiffness matrix.

We decompose the stiffness into a generalized force direction and its orthogonal subspace. Specifically, we assign a high stiffness $k_{high}$ to the directions that are orthogonal to the force. For the force direction, we define an adaptive stiffness scalar $k_{adp}$ that varies with the force magnitude $\|f_{ext}\|$:
\begin{equation}
\small
\label{eq:adaptive_stiffness}
k_{adp} = 
\begin{cases} 
k_{max}, & \|f_{ext}\| < f_{min} \\
k_{max} - \frac{k_{max}-k_{min}}{f_{max}-f_{min}}(\|f_{ext}\| - f_{min}), & \text{otherwise} \\
k_{min}, & \|f_{ext}\| > f_{max}
\end{cases}
\end{equation}
where $f_{min}$ and $f_{max}$ are thresholds determining the sensitivity to contact, and $k_{max}, k_{min}$ define the stiffness range.

\subsubsection{Unified Training Objective}
To incorporate this regularization into the diffusion framework, we use a unified prediction space. We construct an augmented action vector $a_{aug,t}$ by concatenating the original action $a_{t}$, the calculated virtual target $x_{vt}$, and the stiffness magnitude $k_{adp}$:
\begin{equation}
a_{aug,t} = \text{concat}([a_t, x_{vt}, k_{adp}]).
\end{equation}
The diffusion policy is then trained to denoise the sequence $\textbf{A}^0_t\triangleq\{a_{aug, t-h_o+1}, \dots, a_{aug, t-h_o+h_a}\}$ of the augmented vectors.
During inference, we discard the auxiliary components and execute only $\hat{a}_t$.

\subsubsection{Advantages over Force Prediction}
While equation \ref{eq:vt_calculation} implies that predicting $x_{vt}$ is mathematically equivalent to predicting $f_{ext}$ (given $x_{real}$ and $K$), the virtual target offers two significant advantages for representation learning.

The first one is \textbf{objective alignment.}
Raw force prediction uses a sensor-frame signal, whereas the virtual target $x_{vt}$ is computed in the same world-frame Cartesian coordinate system as the action space.
This alignment helps the network to learn a consistent representation for both motion planning and force understanding.

The second one is \textbf{adaptive importance weighting.}
The use of adaptive stiffness acts as a dynamic weighting mechanism.
Consider the deviation $\Delta x \triangleq x_{vt} - x_{real} = K^{-1}f_{ext}$.
\begin{itemize}
    \item In free motion, $\|f_{ext}\|$ is small (mostly sensor noise). According to Eq. \ref{eq:adaptive_stiffness}, $K$ becomes large ($k_{max}$), making $K^{-1}$ small. Consequently, $\Delta x \to 0$, and $x_{vt} \approx x_{real}$.
    \item During contact, $\|f_{ext}\|$ is large. $K$ becomes small ($k_{min}$), making $K^{-1}$ large. This amplifies $\Delta x$, causing $x_{vt}$ to deviate significantly from $x_{real}$. 
\end{itemize}
This mechanism makes high-force contact events contribute more strongly while ignoring noise during free motion, without relying on task-specific manual loss weights.

\subsection{Implementation Details}
\label{sec:implementation}

Performing contact-rich tasks with force control imposes stringent requirements on the execution precision of the entire system.
To ensure better performance, we adjust various system components, ranging from learning objectives to hardware and low-level controller.

\subsubsection{Learning Stability}
\label{sec:learning_stability}
Directly utilizing force signals in an end-to-end network can lead to instability.
We observe that the powerful fitting capability of Transformer-based DP often results in overfitting to high-frequency noise within force signals,
causing action jitter during inference.
We address this through two key modifications.

First, we replace the standard $\epsilon$-prediction parameterization with velocity-prediction.
While $\epsilon$-prediction and sample-prediction are common, we found that velocity-prediction strikes a better balance between inference stability and adherence to conditional information.
Following the formulation in \cite{progressive_diffusion_distillation}, the relationship between velocity $\mathbf{v}_k$, noise $\epsilon$, and the original sample $\mathbf{A}^0_t$ is defined as
\begin{equation}
\label{eq:velocity}
\mathbf{v}^k_t \triangleq \sqrt{\bar{\alpha}_k}\epsilon - \sqrt{1-\bar{\alpha}_k}\mathbf{A}^0_t .
\end{equation}
The corresponding training loss is formulated in \cref{eq:implicitrdp_loss}.
\begin{equation}
\label{eq:implicitrdp_loss}
\mathcal{L}_{\mathbf{v}} = \mathbb{E}_{k, \epsilon^k, (\mathbf{O}_t, \mathbf{A}^0_t)} [\| \mathbf{v}^k_t - \mathbf{v}_\theta(\mathbf{O}_t, \mathbf{A}^k_t, k) \|^2].
\end{equation}

Second, we adopt Euler angles for rotation representation instead of 6D rotation \cite{6d_rotation} or quaternions.
Since the three dimensions of Euler angles are independent, this representation reduces the coupling in rotation regression, thereby further enhancing action stability.
Notably, because our policy predicts relative actions \cite{umi}, the discontinuities and Gimbal lock issues are naturally avoided.

\subsubsection{Hardware Design}
Effective force-based learning requires distinctive physical signals.
When both the end-effector and the manipulated object are rigid,
the variations in action adjustments resulting from force feedback are often subtle and easily drowned out by noise,
significantly increasing the difficulty of policy learning.
To mitigate this, we design a custom compliant fingertip.
This hardware compliance ensures that contact with objects of any stiffness always produces distinctive reactivity signals,
providing the network with clear, high-quality pairs of force feedback and action adjustments to learn from.

\subsubsection{Controller Tuning}
Since ImplicitRDP relies on the policy to learn reactive behaviors based on force,
the low-level controller must provide precise position tracking rather than inherent compliance.
Therefore, for the robot's impedance controller, we tune the integral gain $k_i$ of the PI controller in the Cartesian space by increasing it from $0$ to $10$ while keeping the proportional gain $k_p$ at its default value.
This adjustment ensures that the robot faithfully tracks the action-rate adjustments commanded by the policy, with approximately $0.1$~mm position tracking precision after convergence in our setup.

%% file: sections/4_experiments.tex
% !TEX root = ../main.tex
\section{Experiments}

We evaluate ImplicitRDP on real-world contact-rich manipulation tasks to answer four key questions:
\begin{itemize}
    \item \textbf{Q1:} How does the end-to-end closed-loop network compare against visual-only (DP) and hierarchical visual-force (RDP) baselines?
    \item \textbf{Q2:} Does the closed-loop force control in structural slow-fast learning (SSL) improve performance in contact-rich tasks?
    \item \textbf{Q3:} How effectively does the virtual-target-based representation regularization (VRR) compare to other auxiliary tasks?
    \item \textbf{Q4:} Does the velocity-prediction parameterization and rotation representation improve learning stability?
\end{itemize}

\subsection{Experimental Setup}
\label{sec:experimental_setup}

\begin{figure}[htbp]
    \centering
    \includegraphics[width=0.8\linewidth]{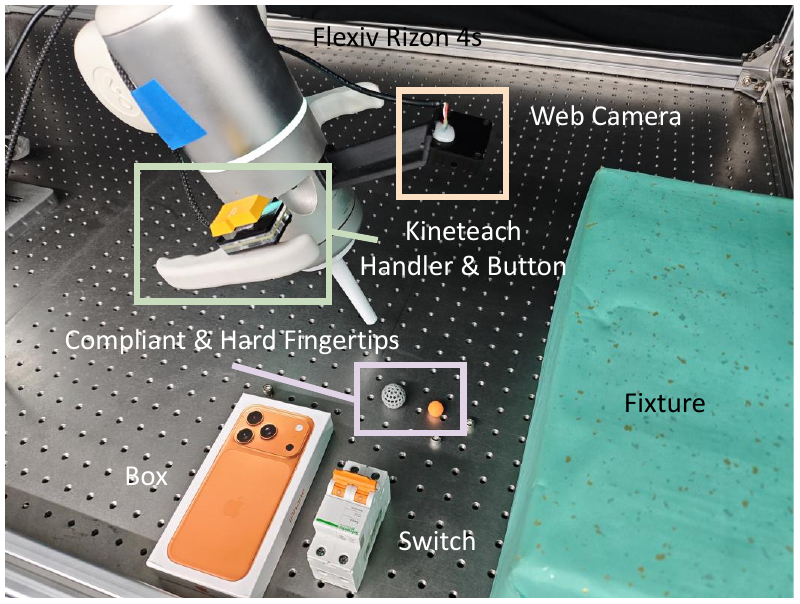}
    \vspace{-4mm}
    \caption{\textbf{Hardware Setup.}
    The system utilizes a Flexiv Rizon 4s robot arm.
    A handle and button are mounted between the seventh joint and 6-axis F/T sensor for kinematic teaching.
    We also design a custom compliant fingertip to ensure distinctive reactivity signals during contact-rich interactions.}
    \vspace{-4mm}
    \label{fig:hardware_setup}
\end{figure}

Our hardware setup (see \cref{fig:hardware_setup}) is built on a Flexiv Rizon 4s \cite{flexiv_rizon4} robot arm, which is equipped with both joint torque sensors and a 6-axis force/torque sensor at the end effector.
This allows us to use the joint torque sensors for kinematic teaching during data collection,
while obtaining contact forces directly from the 6-axis F/T sensor at the same time.
A joystick and our custom compliant fingertips are mounted on the robot’s end effector.
We use a webcam as a wrist camera to record visual observations.

During data collection, camera observations and the policy-used force/torque stream are both recorded at $10$~Hz.
The Flexiv Rizon 4s internal F/T sensor delay is below $20$~ms, and the multi-sensor synchronization latency in our ROS2 system is below $120$~ms.

During training, we set $f_{min}=0.5~\mathrm{N}$, $f_{max}=5~\mathrm{N}$, $k_{min}=200~\mathrm{N/m}$, and $k_{max}=10000~\mathrm{N/m}$ for the VRR adaptive-stiffness schedule in all experiments.
All augmented target dimensions, including the action, virtual target, and adaptive stiffness scalar, are min-max normalized before diffusion training, and unit loss weights are used for these components across all tasks.

During policy execution, the slow observation encoder inference latency and other system execution delays are handled by the consistent inference mechanism in Alg.~\ref{alg:implicitrdp_inference}.
The fast closed loop also updates at $10$~Hz, while the slow loop refreshes every $0.6$~s.
On our inference machine with an Intel Core i9-13900K CPU and an NVIDIA GeForce RTX 4090 GPU, the inference latency per fast-loop step is below $30$~ms, which is within the $100$~ms control budget, and we observed no missed-deadline cases.

\begin{figure*}[htbp]
    \centering
    \includegraphics[width=0.95\linewidth]{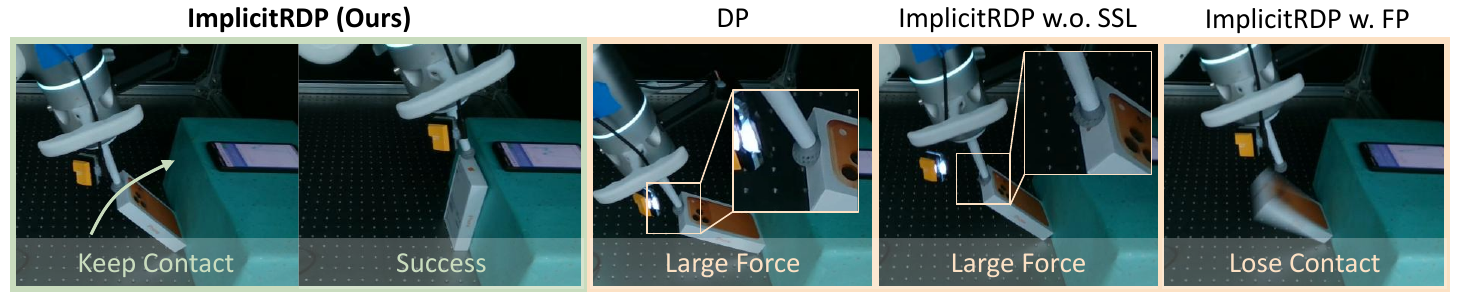}
    \vspace{-4mm}
    \caption{\textbf{Box Flipping Task and Failure Cases.}
    The goal is to push a thin phone box against a fixture to flip it upright while maintaining a delicate force limit ($<14N$).
    Vision-only or open-loop baselines lack closed-loop, force-based adjustment and apply excessive force, resulting in squeezing the fingertip.
    ImplicitRDP successfully utilizes the force feedback to complete the task safely.}
    \label{fig:box_flipping}
    \vspace{-4mm}
\end{figure*}

\begin{figure*}[htbp]
    \centering
    \includegraphics[width=0.95\linewidth]{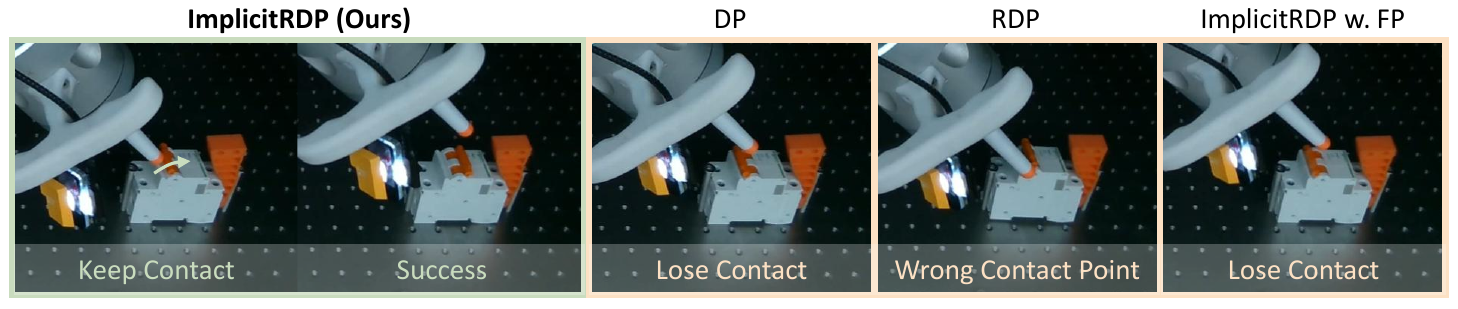}
    \vspace{-4mm}
    \caption{\textbf{Switch Toggling Task and Failure Cases.}
    The robot has to locate and apply a specific force to toggle a circuit breaker switch.
    DP tends to initiate the toggling motion prematurely before the triggering force threshold is reached, while RDP often misses the precise contact location due to latent compression errors.
    ImplicitRDP accurately approaches the switch and perceives force to toggle the switch successfully.}
    \vspace{-4mm}
    \label{fig:switch_toggling}
\end{figure*}

\begin{figure*}[htbp]
    \centering
    \includegraphics[width=0.95\linewidth]{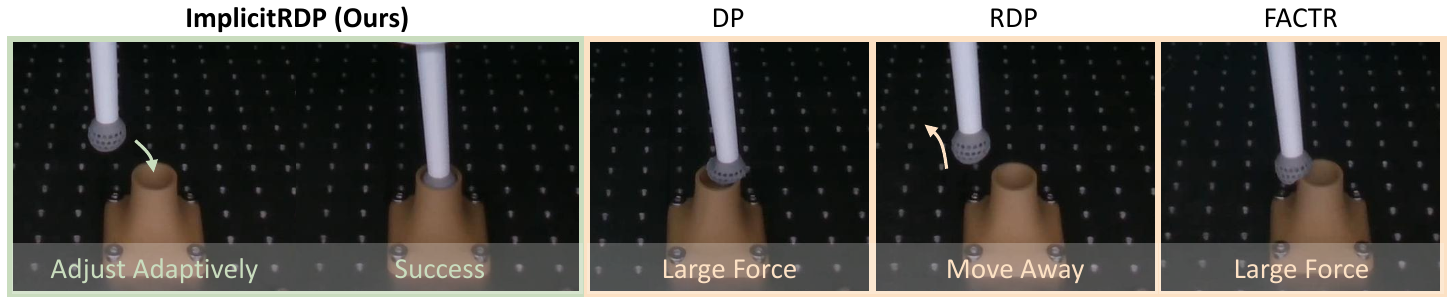}
    \vspace{-4mm}
    \caption{\textbf{Hole Inserting Task and Failure Cases.}
    The robot must insert the compliant fingertip into a tight hole with only a $1$~mm margin.
    DP and FACTR have limited ability to correct the fingertip pose after contacting the base at an incorrect location, while RDP can drift away from the hole due to latent action errors under visually out-of-distribution states.
    ImplicitRDP keeps contact-guided correction and aligns the fingertip before insertion.}
    \vspace{-4mm}
    \label{fig:hole_inserting}
\end{figure*}

We design three representative contact-rich manipulation tasks:
\begin{enumerate}
    \item \textbf{Box Flipping:}
    The robot must push a thin phone box against a fixture to flip it from a flat position to an upright one (see \cref{fig:box_flipping}).
    To increase the difficulty, we intentionally applied a relatively small contact force ($\sim8N$) during demonstration collection.
    During evaluation, any case where the applied force exceeds $14N$ is considered a failure.
    This prevents the policy from completing the task through brute-force pushing.
    This task represents a class of behaviors that require sustained force application, such as fruit peeling and vase wiping.
    \item \textbf{Switch Toggling:}
    The robot needs to toggle a circuit breaker switch (see \cref{fig:switch_toggling}).
    The challenge of this task is that the switch requires a relatively large force to actuate,
    while vision-only policy cannot determine whether the triggering threshold has been reached.
    Unlike box flipping, switch toggling requires a short-duration force burst,
    which is common in tasks like vegetable chopping.
    \item \textbf{Hole Inserting:}
    The robot must insert the compliant fingertip into a hole whose diameter leaves only a $1$~mm margin (see \cref{fig:hole_inserting}).
    This task requires the policy to use contact feedback to correct visual localization error during insertion, which is common in peg-in-hole and other assembly tasks.
\end{enumerate}
For each task, we collected 40 demonstrations.
Each reported success rate is computed from $20$ binary success/failure trials.

\subsection{Baselines}
We compare ImplicitRDP against the following baselines:
\begin{itemize}
    \item \textbf{Diffusion Policy (DP):}
    Standard CNN-based DP with vision-based open-loop control.
    \item \textbf{Reactive Diffusion Policy (RDP):}
    The state-of-the-art hierarchical slow-fast visual-force learning method.
    \item \textbf{FACTR:}
    Transformer DP with FACTR's blurred-vision curriculum.
    \item \textbf{FACTR + SSL:}
    Transformer DP with both SSL and FACTR's blurred-vision curriculum.
    \item \textbf{ImplicitRDP w.o. SSL and VRR:}
    Similar to standard transformer-based DP, but augmented with open-loop force inputs and techniques that improve training stability in \cref{sec:learning_stability}.
    \item \textbf{ImplicitRDP w.o. SSL:}
    Similar to the previous one, but with VRR used.
    \item \textbf{ImplicitRDP w. Different Auxiliary Tasks:}
    Use no other task or use force prediction as the auxiliary task.
    \item \textbf{ImplicitRDP w. Different Training Choices:}
    Use alternative implementation choices discussed in \cref{sec:learning_stability}.
\end{itemize}
We keep the slow visual update schedule consistent across all models.

\begin{figure}[htbp]
    \centering
    \includegraphics[width=0.8\linewidth]{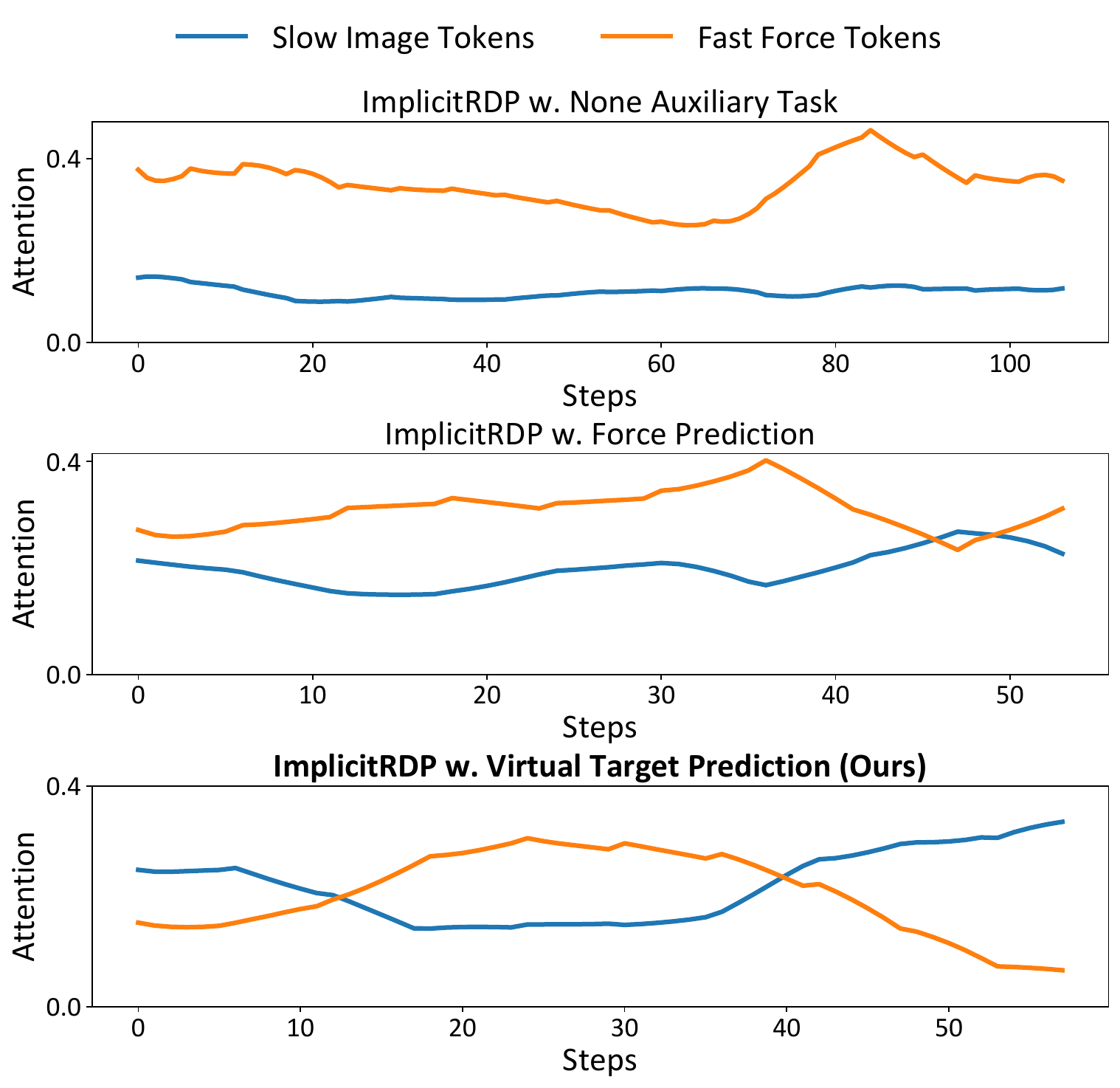}
    \vspace{-4mm}
    \caption{\textbf{Attention Weight Visualization.}
    We visualize the summed attention weights of visual tokens and force tokens from the first transformer layer in the switch toggling task.
    The weights are averaged over all heads and all queries.
    A sliding window of size 10 is applied to smooth the curves.}
    \vspace{-4mm}
    \label{fig:attention}
\end{figure}

\subsection{Results and Analysis}

\subsubsection{Comparison with Baselines (Q1)}
As illustrated in \cref{tab:main_results}, the end-to-end ImplicitRDP consistently achieves the best performance compared to both the vision-only DP and the hierarchical visual-force policy RDP.

Failure cases in \cref{fig:box_flipping} reveal that in the box flipping task, the vision-only DP often applies force far exceeding normal levels, resulting in dangerous crushing of the phone box.
This failure stems from the inability of visual observations to determine whether the applied force is appropriate.
Similarly, in the switch toggling task, as shown in \cref{fig:switch_toggling}, DP tends to start the upward toggling motion before the required triggering force is reached, as the visual difference between the triggered and un-triggered states is negligible.
The same limitation appears in Hole Inserting (see Fig.~\ref{fig:hole_inserting}), where DP cannot correct the fingertip pose after an inaccurate initial contact with the base.

Regarding RDP, while it performs adequately in box flipping, it struggles with switch toggling.
\cref{fig:switch_toggling} shows that RDP frequently contacts the wrong location during the approach phase.
We hypothesize that this is because the fast policy in RDP compresses raw actions into a latent space, leading to precision loss during free-space motion.
In Hole Inserting, RDP can also drift away from the hole due to latent action errors under visually out-of-distribution states, as illustrated in Fig.~\ref{fig:hole_inserting}.

We further evaluate FACTR-style blurred-vision curriculum baselines on Hole Inserting.
FACTR and FACTR + SSL achieve $4/20$ and $1/20$ success respectively.
These results show that both the open-loop and closed-loop variants fail to achieve the force-adaptive adjustment enabled by VRR, indicating that the blurred-vision curriculum is not as effective as VRR for contact-guided correction.

In contrast, ImplicitRDP realizes closed-loop control based on force signals and performs end-to-end denoising directly in the original action space.
This allows it to adaptively weigh different modalities, ensuring accurate reactivity during contact, precise movement during free motion, and contact-guided correction.

\begin{table}[htbp]
\caption{Success Rate Compared with Baseline Methods}
\vspace{-4mm}
\label{tab:main_results}
\begin{center}
\resizebox{\linewidth}{!}{%
\begin{tabular}{lccc}
\toprule
\textbf{Method} & \textbf{Box Flipping} & \textbf{Switch Toggling} & \textbf{Hole Inserting} \\
\midrule
DP & 0/20 & 8/20 & 0/20 \\
RDP & 16/20 & 10/20 & 5/20 \\
\textbf{ImplicitRDP (Ours)} & \textbf{18/20} & \textbf{18/20} & \textbf{8/20} \\
\bottomrule
\end{tabular}
}
\end{center}
\vspace{-4mm}
\end{table}

\subsubsection{Effectiveness of Closed-Loop Control (Q2)}
To validate the importance of the closed-loop mechanism provided by Structural Slow-Fast Learning (SSL), we compare the full model against open-loop variants.
As shown in \cref{tab:closed_loop}, when both SSL and Virtual-target-based Representation Regularization (VRR) are removed, and the network relies only on low-frequency visual and force signals for open-loop control, performance drops significantly across both tasks.
Even when VRR is reintroduced, the open-loop variant still suffers from a performance decline compared to the complete ImplicitRDP.

Notably, the performance drop is much more pronounced in the box flipping task.
According to \cref{fig:box_flipping}, the primary cause of failure in the open-loop setting is excessive force application.
This is probably because box flipping requires the sustained application of a constant force, while an open-loop network cannot adjust its actions at the action rate within a chunk based on force feedback.
Consequently, the applied force deviates from the target, pushing the state into an out-of-distribution region and leading to task failure.
These results demonstrate that the closed-loop force control realized by SSL in ImplicitRDP is critical for improving performance in contact-rich tasks, particularly those requiring sustained force maintenance.

\begin{table}[htbp]
\caption{Comparison between Open-loop and Closed-loop Control}
\vspace{-4mm}
\label{tab:closed_loop}
\begin{center}
\begin{tabular}{lcc}
\toprule
\textbf{Method} & \textbf{Box Flipping} & \textbf{Switch Toggling} \\
\midrule
ImplicitRDP w.o. SSL and VRR & 6/20 & 5/20 \\
ImplicitRDP w.o. SSL & 4/20 & 15/20 \\
\textbf{ImplicitRDP (Ours)} & \textbf{18/20} & \textbf{18/20} \\
\bottomrule
\end{tabular}
\end{center}
\vspace{-4mm}
\end{table}

\subsubsection{Auxiliary Task Analysis (Q3)}
We further analyze the impact of different auxiliary tasks in \cref{tab:auxiliary_tasks}.
We find that using the virtual target as the prediction objective yields the best performance on both tasks.
While standard force prediction provides some improvement over using no auxiliary task, it remains inferior to VRR.
Failure cases from \cref{fig:box_flipping} and \cref{fig:switch_toggling} illustrates that policies trained with other auxiliary strategies tend to lose contact untimely during both box pushing and switch toggling.
\cref{fig:attention} shows that without the auxiliary task, the model fails to learn the importance relationships between different modalities.
These results confirm that these networks fail to fully utilize action-rate force inputs, resulting in modality collapse.
In comparison, VRR in ImplicitRDP resides in the same space as the action and employs adaptive weighting,
which helps regularize the model representation, encouraging the network to focus on critical force data and thereby enhancing performance.

\begin{table}[htbp]
\caption{Comparison of Different Auxiliary Tasks}
\vspace{-4mm}
\label{tab:auxiliary_tasks}
\begin{center}
\begin{tabular}{lcc}
\toprule
\textbf{Auxiliary Task} & \textbf{Box Flipping} & \textbf{Switch Toggling} \\
\midrule
None & 6/20 & 6/20 \\
Force Prediction & 8/20 & 10/20 \\
\textbf{Virtual Target Prediction (Ours)} & \textbf{18/20} & \textbf{18/20} \\
\bottomrule
\end{tabular}
\end{center}
\vspace{-4mm}
\end{table}

\subsubsection{Ablation on Learning Stability (Q4)}
\cref{tab:ablation} shows the impact of prediction parameterization and rotation representation.
Results indicate that our choice of velocity-prediction consistently outperforms $\epsilon$-prediction and sample-prediction,
particularly in the box flipping task which requires continuous force application.
Furthermore, using Euler angles proves superior to 6D rotation.
The latter struggles with unstable actions in switch toggling due to worse noise tolerance caused by non-independent representation.
In general, the combination of velocity-prediction and Euler angles achieves the highest stability and success rates across both tasks.

\begin{table}[htbp]
\caption{Ablation Study on Learning Stability}
\vspace{-5mm}
\label{tab:ablation}
\begin{center}
\begin{tabular}{lcc}
\toprule
\textbf{Method} & \textbf{Box Flipping} & \textbf{Switch Toggling} \\
\midrule
ImplicitRDP ($\epsilon$-prediction) & 9/20 & 18/20 \\
ImplicitRDP (sample-prediction) & 7/20 & 14/20 \\
ImplicitRDP (6D Rotation) & 16/20 & 12/20 \\
\textbf{ImplicitRDP (Ours)} & \textbf{18/20} & \textbf{18/20} \\
\bottomrule
\end{tabular}
\end{center}
\vspace{-5mm}
\end{table}

%% file: sections/5_conclusion.tex
% !TEX root = ../main.tex
\section{Conclusion}

In this paper, we present ImplicitRDP, a novel end-to-end framework that reconciles low-frequency visual planning and rapid force control at the action rate.
By embedding structural slow-fast learning directly into the diffusion process, we eliminate the need for separate policy hierarchies,
allowing a single network to dynamically attend to different modalities with variant frequencies.
Additionally, our proposed virtual-target auxiliary task effectively regularizes the representation space, 
ensuring the policy adaptively leverages physical feedback rather than over-relying on one single modality.
Experimental results confirm that this unified approach not only simplifies the training pipeline but also achieves superior performance in contact-rich manipulation compared with baselines. 
Future work will investigate extending this unified framework to Vision-Language-Action (VLA) models, as well as integrating higher-rate F/T sensors or other fast sensing modalities such as tactile sensing.